\documentclass[runningheads]{llncs}
\usepackage{booktabs}
\usepackage{amsmath}
\usepackage[T1]{fontenc}
\usepackage{graphicx,verbatim}

\usepackage{subcaption}
\usepackage{caption}
\usepackage{float}
\usepackage{xcolor}
\usepackage{hyperref}

\hypersetup{colorlinks=true,     breaklinks=true,              %
linkcolor=blue,  
urlcolor=blue,       %
citecolor=blue}
\begin{document}
\title{
\texttt{\textbf{Mammo-SAE}}: %
Interpreting Breast Cancer Concept Learning with Sparse Autoencoders
}
\author{Krishna Kanth Nakka%
}

\authorrunning{Krishna Kanth Nakka}

\institute{Bavaria, Germany \\
\email{krishkanth.92@gmail.com}\\
\url{https://krishnakanthnakka.github.io} 
}

\maketitle              %

\begin{abstract}
Interpretability is critical in high-stakes domains such as medical imaging, where understanding model decisions is essential for clinical adoption. In this work, we introduce Sparse Autoencoder (SAE)-based interpretability to breast imaging by analyzing {Mammo-CLIP}, a vision--language foundation model pretrained on large-scale mammogram image--radiology report pairs. We train a patch-level \texttt{Mammo-SAE} on Mammo-CLIP visual features to identify and probe latent neurons associated with clinically relevant breast concepts such as \textit{mass} and \textit{suspicious calcification}. We show that top-activated class-level latent neurons often tend to align with ground-truth regions, and also uncover several confounding factors influencing the model’s decision-making process. Furthermore, we demonstrate that finetuning Mammo-CLIP leads to larger concept separation in the latent space, improving interpretability and predictive performance. Our findings suggest that sparse latent representations offer a powerful lens into the internal behavior of breast foundation models.  The code will be released at \url{https://krishnakanthnakka.github.io/MammoSAE/}.
\end{abstract}

\section{Introduction}

In high-stakes domains such as healthcare, machine learning models must not only be accurate but also interpretable. To enhance transparency in breast imaging, prior work has proposed both post-hoc interpretability methods—such as GradCAM variants~\cite{raghavan2024attention,liu2024breast,kajala2024breaking}—and inherently interpretable architectures, including those based on prototype networks~\cite{choukali2024pseudo,pathak2024prototype}. While these tools provide useful explanations at the prediction level, they offer limited insight into the model’s internal mechanisms, particularly at the level of individual neurons. Moreover, prior studies have shown that neurons in deep networks are often \textit{polysemantic}~\cite{o2023disentangling,dreyer2024pure,marshall2024understanding}, i.e., they activate in response to multiple unrelated concepts, making them difficult to interpret reliably.

Recently, Sparse Autoencoders (SAEs)~\cite{cunningham2023sparse,gao2024scaling,makelov2024towards,lieberum2024gemma,he2024llama} have gained significant traction for interpreting Large Language Models (LLMs)~\cite{team2024gemma,touvron2023llama}. Building on this progress,  SAEs have also been adapted to Vision Language Models~\cite{rao2024discover,lim2024sparse}. SAEs are capable of extracting \textit{monosemantic} features—latent dimensions that correspond to interpretable concepts—and support test-time interventions that allow controlled probing and manipulation of model behavior. Notably, SAEs can be integrated at any layer of a model, providing a flexible and modular approach to interpreting intermediate representations. This layer-wise adaptability makes them a powerful tool for dissecting model behavior at inference time. 

\vspace{0.5em}
In this work, we extend SAE-based interpretability techniques to breast imaging by introducing \texttt{Mammo-SAE}, a sparse autoencoder trained on visual features from {Mammo-CLIP}~\cite{ghosh2024mammo}, a vision--language foundation model pretrained on mammogram image--report pairs. Our contributions are as follows:
\begin{enumerate}
\item We train a patch-based SAE on Mammo-CLIP to discover latent neurons associated with breast cancer-related concepts such as \textit{mass} and \textit{suspicious calcification}.
\item We uncover monosemantic features in the SAE latent space and visualize their spatial activation patterns, showing alignment with clinical regions of interest.
\item We conduct targeted group interventions on the SAE latent space to reveal that the model sometimes relies on confounding features when making decisions. 
\item We find that finetuning leads to a clearer separation of latent neurons associated with breast cancer-related concepts, providing insight into observed performance gains.

\end{enumerate}

\begin{figure}[t]
\centering
\includegraphics[width=0.75\linewidth]{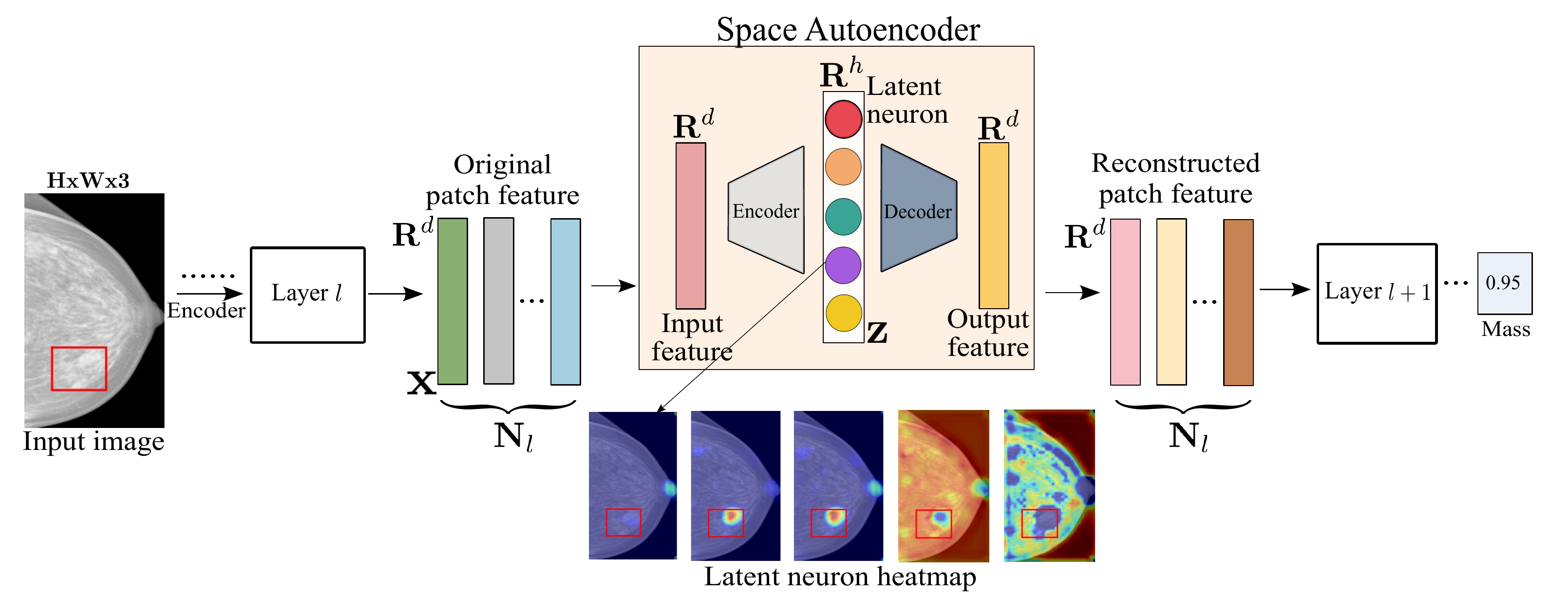}
\caption{{\bf Mammo-SAE Framework.} The SAE is first trained on patch-level CLIP features  $\mathbf{x}_j \in \mathbf{R}^d$ at any given layer, projecting them into a high-dimensional, interpretable sparse latent space  $\mathbf{z} \in \mathbf{R}^h$, and decoding them back for reconstruction. Once trained, the SAE is used to analyze which latent neurons are activated and what semantic information they encode. We also perform targeted interventions in the latent neuron space to assess their influence on downstream label prediction. We observe the learned latents capture diverse regions such as \textit{nipple regions, masses}, and background areas. Red box indicate ground-truth mass localization.}
\label{fig:framework}
\end{figure}

\section{Proposed Method}

Figure~\ref{fig:framework} illustrates our proposed \texttt{\textbf{Mammo-SAE}} framework for interpreting breast foundation models; in this work, we apply it to {Mammo-CLIP} as a representative example. The framework consists of three main components: (i) an encoder-decoder SAE is pretrained to project CLIP features into a high-dimensional sparse latent space, encouraging disentangled and interpretable representations (Sec~\ref{sec:SAE}); (ii) a probing framework is employed to identify latent neurons that are selectively activated in the presence of breast cancer-related concepts such as \textit{mass} and \textit{suspicious calcification}, enabling concept-level interpretability (Sec~\ref{sec:saescore}); and (iii) an intervention framework is used to manipulate group of class-level latent neurons and observe changes in the model’s output, allowing us to assess the causal impact of latent neurons and identify potential confounding factors influencing decision-making (Sec~\ref{sec:intervention}).\\

\noindent \textbf{Preliminaries.} {Mammo-CLIP}~\cite{ghosh2024mammo} is a vision--language foundation model trained to align image and text representations using paired mammogram images and radiology reports. After pretraining, the Mammo-CLIP image embeddings can be used for downstream concept prediction (e.g., binary classification of the presence of a \textit{mass}) via a single fully connected classification layer. Additionally, the model can be further finetuned by updating the CLIP backbone and classifier jointly to improve performance on specific breast concept recognition tasks. Throughout this paper, we refer to the original frozen backbone as the \textit{pretrained} variant and the end-to-end updated model as the \textit{finetuned} variant. We provide further details in Section~\ref{sec:additionaldetails} of the Appendix.

\subsection{\texttt{Mammo-SAE}}\label{sec:SAE}

Let an input image $I$ and the feature extracted by the Mammo-CLIP model $f$ at layer $l$ is $\mathbf{x}$. SAE takes the Mammo-CLIP local feature $\mathbf{x}_l^j \in \mathbf{R}^d$ at layer $l$ and spatial position $j$, where $j$ indexes the spatial location in the feature map of size $N_l = H_l \times W_l$. The SAE consists of two layers: an encoder that projects the input into a high-dimensional sparse latent space, and a decoder that reconstructs the original CLIP feature from this latent representation.

The model is trained using a combination of reconstruction loss and a sparsity constraint, which encourages activation of only a small subset of neurons in the latent space, thereby enhancing interpretability. Let $W_{\mathrm{enc}} \in \mathbf{R}^{d \times h}$ and $W_{\mathrm{dec}} \in \mathbf{R}^{h \times d}$ denote the encoder and decoder weight matrices, respectively, and let $\phi(\cdot)$ denote the ReLU non-linear function. The training objective is defined as:

\begin{equation}
    \mathcal{L} = 
    \underbrace{\left\| W_{\mathrm{dec}} \, \phi\left(W_{\mathrm{enc}} \mathbf{x}^j \right) - \mathbf{x}^j \right\|_2^2}_{\mathrm{Reconstruction  Loss}} 
    + 
    \lambda \underbrace{\left\| \phi\left(W_{\mathrm{enc}} \mathbf{x}^j \right)\right\|_1}_{\mathrm{Sparsity Loss}},
\end{equation}

where the first term represents the reconstruction loss, and the second term is the sparsity loss with regularization coefficient $\lambda$. Further implementaton details about the SAE training can be found in Section~\ref{sec:exps}.

\subsection{Probing Mammo-SAE Latents to Identify Breast Concepts}\label{sec:saescore}

After training Mammo-SAE on layer-$l$ activations from Mammo-CLIP, we analyze the resulting latent space to identify neurons that correspond to specific breast cancer-related concepts in the model. Our goal is to pinpoint latent neurons that are consistently activated in the presence of concepts such as \textit{mass} or \textit{suspicious calcification}.

Given a dataset of $\{\mathbf{x}, \mathbf{c}\}$ where each image $\mathbf{x}$ is annotated with a class label $c \in \{0, 1\}$, where $c = 1$ denotes the presence of the breast concept (eg., mass), we compute the class-wise mean latent activation vector $\bar{\mathbf{z}}^{(c)} \in \mathbf{R}^h$ by averaging over all spatial locations and all training samples in that class:

\begin{equation}
\bar{\mathbf{z}}^{(c)} = \frac{1}{|\mathcal{D}_c| \cdot N_l} \sum_{\mathbf{x} \in \mathcal{D}_c} \sum_{j=1}^{N_l} \phi\left(W_{\mathrm{enc}}\, \mathbf{x}^j\right),
\end{equation}

where $\mathcal{D}_c$ is the set of training images with class label $c$, and $\mathbf{x}^j \in \mathbf{R}^d$ denotes the CLIP feature at spatial location $j$ in image feature $\mathbf{x}$. The function $\phi(\cdot)$ denotes the ReLU activation applied after encoding.

We then assign each latent neuron \( t \) a class-level relevance score defined as its class-specific mean activation, \( s_t^{(c)} = \bar{z}_t^{(c)} \), where \( \bar{z}_t^{(c)} \) is the \( t \)-th element of \( \bar{\mathbf{z}}^{(c)} \). Latent neurons are ranked in descending order of \( s_t^{(c)} \), and the top-scoring ones for class \( c = 1 \) are considered most aligned with the target concept.  While we adopt a simple mean-based scoring approach, alternative strategies based on entropy or standard deviation can also be explored~\cite{lim2024sparse}. To assess the reliability of the top class-level latent neurons, we examine the image regions that most strongly activate each neuron. This analysis provides visual evidence of whether a neuron attends to clinically meaningful regions or to spurious patterns, offering insight into the model’s internal reasoning.

\subsection{Intervention on Mammo-SAE Latent Neurons}\label{sec:intervention}

Furthermore, to assess the causal role of SAE latent neurons in downstream predictions, we perform targeted interventions on the top-$k$ class-level neurons identified for a given concept. Specifically, we introduce two types of group interventions on the patch-level SAE latent $\mathbf{z} = \phi\left(W_{\mathrm{enc}}\, \mathbf{x}^j\right)$ at every spatial position $j$:

(i) \textbf{Top-$k$ Activated:} We retain only the activations of the top-$k$ latent neurons and zero out all others. This isolates the influence of the most concept-relevant neurons.

\begin{equation}
\mathbf{z}' = \mathbf{z} \odot \mathbf{m}, \quad \text{where } m_i =
\begin{cases}
1, & \text{if } i \in \mathcal{T}_k^{(0)} \cup \mathcal{T}_k^{(1)} \\
0, & \text{otherwise}
\end{cases}
\label{eq:topk-activated}
\end{equation}

Here, \( \mathcal{T}_k^{(c)} \subset \{1, \dots, k\} \) denotes the set of indices corresponding to the top-\(k\) neurons for class \( c \in \{0, 1\} \), ranked by their class-specific scores \( s_k^{(c)} \). A neuron position \( i \) is retained (i.e., \( m_i = 1 \)) if it appears in the union of top-\(k\) neurons for class 0 or class 1.

(ii) \textbf{Top-$k$ Deactivated:} We zero out the top-$k$ neurons while leaving all other latent activations unchanged. This tests the dependency of the model's prediction on these specific neurons.

\begin{equation}
\mathbf{z}' = \mathbf{z} \odot (1 - \mathbf{m})
\label{eq:topk-deactivated}
\end{equation}

By measuring the change in the model’s output before and after these interventions, we assess the functional importance of the selected neurons and determine whether the model relies on meaningful features or potentially confounding patterns.

\section{Experiments}\label{sec:exps}

\noindent \textbf{Dataset.} We conduct our experiments on the VinDr-Mammo dataset~\cite{nguyen2023vindr}, which contains approximately 20,000 mammogram images from 5,000 patients. Each image is annotated with breast-specific concepts, including the presence of \textit{mass} and \textit{suspicious calcification}.

\vspace{0.5em}
\noindent \textbf{SAE Training.} We utilize the Vision-SAEs library~\cite{rao2024discover} to train a Sparse Autoencoder (SAE) on patch-level features extracted from the fine-tuned Mammo-CLIP model~\cite{ghosh2024mammo}. We focus specifically on the classifier trained for the \textit{suspicious calcification} concept, using \textit{activations from the final layer} of the EfficientNet-B5 backbone~\cite{koonce2021efficientnet} of Mammo-CLIP~\cite{ghosh2024mammo}. Rather than training separate SAEs for each model which is expensive, we train a single SAE once and reuse it across all experiments. This shared SAE design not only reduces overhead but also ensures a consistent latent space, making it easier to compare representations across different models (see Section~\ref{sec:latentseparationsc}).  The input feature dimension is $d = 2048$, and we set the expansion factor to 8, resulting in a latent dimension of $h = 16{,}384$. The SAE is trained for 200 epochs with a learning rate of $3 \times 10^{-4}$, sparsity penalty $\lambda = 3 \times 10^{-5}$, and batch size of 4096. 

\vspace{0.5em}
\noindent \textbf{Metrics.} We follow the evaluation protocol in~\cite{ghosh2024mammo} and report the AUC-ROC for the binary classification tasks at hand.

\vspace{0.5em}
\noindent \textbf{SAE Generalizability.} In Table~\ref{tab:saeperformace}, we compare the predictive performance of models using original CLIP features versus SAE-reconstructed features, for both \textit{mass} and \textit{suspicious calcification} concepts. We conduct this comparison on both the pretrained and fine-tuned variants of Mammo-CLIP to assess whether the SAE preserves original information. Across both models and both concepts, we observe that the drop in AUC-ROC is less than 2\%, indicating that the SAE—trained once on a single network—generalizes well and retains reliable representations for downstream prediction. We will now explore SAEs to dissect the model behaviour.

\begin{table}[t]
\centering
\caption{AUC-ROC comparison between the original Mammo-CLIP model and the SAE-reconstructed model. We insert SAE at the final layer.}
\label{tab:saeperformace}
\begin{subtable}[t]{0.45\textwidth}
\centering
\begin{tabular}{lcc}
\toprule
\textbf{} & {W/o SAE} & {W/ SAE} \\
\midrule
Pretrained & 0.951 & 0.933 \\
Finetuned  & 0.978 & 0.979 \\
\bottomrule
\end{tabular}
\caption{Suspicious calcification}
\end{subtable}
\hfill
\begin{subtable}[t]{0.45\textwidth}
\centering
\begin{tabular}{lcc}
\toprule
\textbf{} & {W/o SAE} & {W/ SAE} \\
\midrule
Pretrained & 0.786 & 0.763 \\
Finetuned  & 0.856 & 0.855 \\
\bottomrule
\end{tabular}
\caption{Mass}
\end{subtable}\vspace*{-0.5cm}
\end{table}

\vspace*{-0.2cm}
\subsection{Intervention on Class-level Latent Neurons}

As described in Section~\ref{sec:saescore}, we compute the relevance score of each latent neuron with respect to two classes (e.g., \textit{Mass} vs. \textit{Non-Mass}) and identify the top-$k$ neurons per class. We then perform targeted interventions as outlined in Section~\ref{sec:intervention}.

\begin{figure}[H]
    \centering
    \begin{subfigure}[t]{0.48\linewidth}
        \centering
        \includegraphics[width=\linewidth]{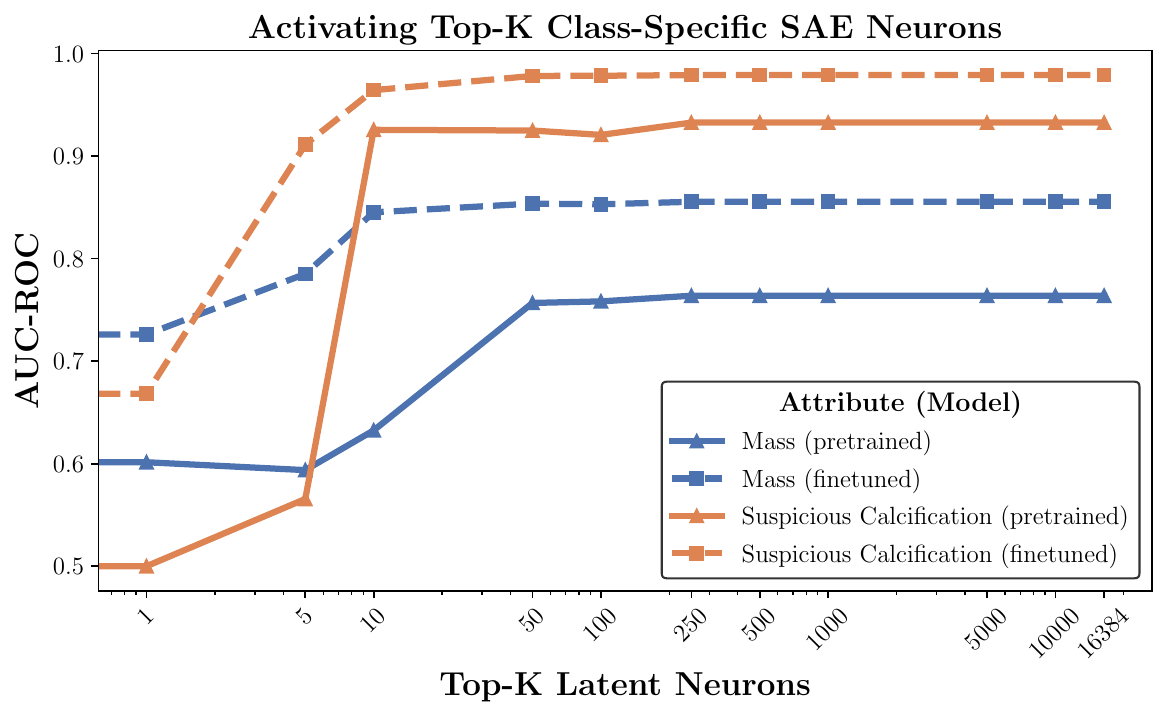}
        \caption{Top-$k$ Activated}
        \label{fig:topkon}
    \end{subfigure}
    \begin{subfigure}[t]{0.48\textwidth}
        \centering
        \includegraphics[width=\linewidth]{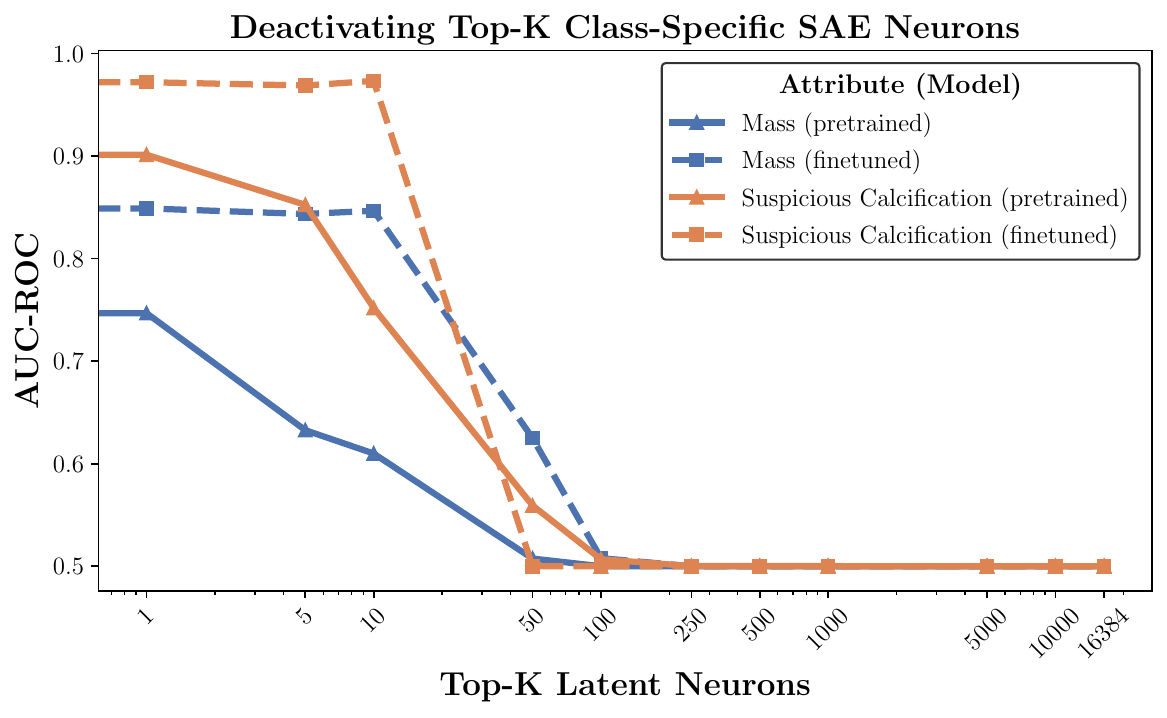}
        \caption{Top-$k$ Deactivated}
        \label{fig:topkoff}
    \end{subfigure}
   \caption{{\bf Intervention on class-level latent neurons.} Left: Top-$k$ activated intervention—only the top-$k$ class-specific neurons are retained, and all others are zeroed out. Right: Top-$k$ deactivated intervention—the top-$k$ neurons are zeroed out while the rest remain unchanged.}
    \label{fig:combined}
\end{figure}

In the left panel of Figure~\ref{fig:topkon}, we show results for the \textbf{top-$k$ activated} intervention, where only the top-$k$ class-specific neurons are retained and all others are zeroed out, with $k$ varied from 0 up to the full latent dimensionality $h = 16{,}384$. Remarkably, activating as few as 10 neurons is sufficient to nearly recover the model's original AUC-ROC in multiple cases—demonstrating that \textbf{a small subset of neurons captures most of the task-relevant signal.}

Conversely, in Figure~\ref{fig:topkoff}, we present results for the \textbf{top-$k$ deactivated} intervention, where the top-$k$ class-specific neurons are zeroed out while the rest of the latent representation is left unchanged. We observe that deactivating more than 10 neurons leads to a sharp drop in AUC-ROC, highlighting the model’s strong reliance on a compact, concept-aligned subset of latent features. These findings underscore the precision and interpretability of the Mammo-SAE representation.

\subsection{Analyzing Top Activated Latent Neurons}

To interpret the internal representations learned by Mammo-SAE, we visualize the activations of the top-$k$ latent neurons from the encoded representation $\mathbf{z}$ at each spatial location $j$.

\begin{figure}[H]
\centering
\begin{subfigure}[t]{0.9\linewidth}
    \centering
    \includegraphics[width=\linewidth]{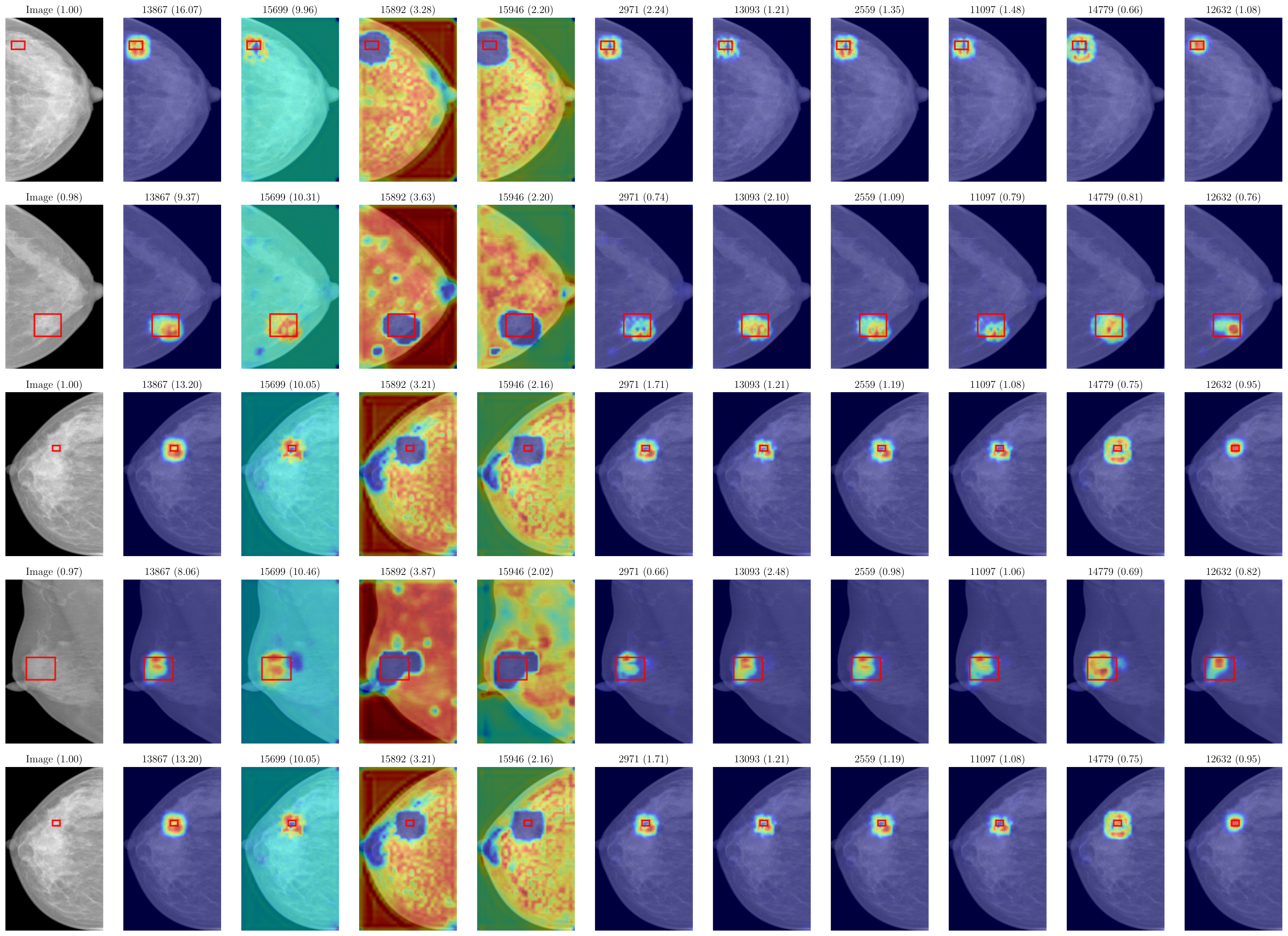}
    \caption{{Suspicious Calcification} (Finetuned)}
    \label{fig:topkactivatedsc}
\end{subfigure}
\hfill
\begin{subfigure}[t]{0.9\linewidth}
    \centering
    \includegraphics[width=\linewidth]{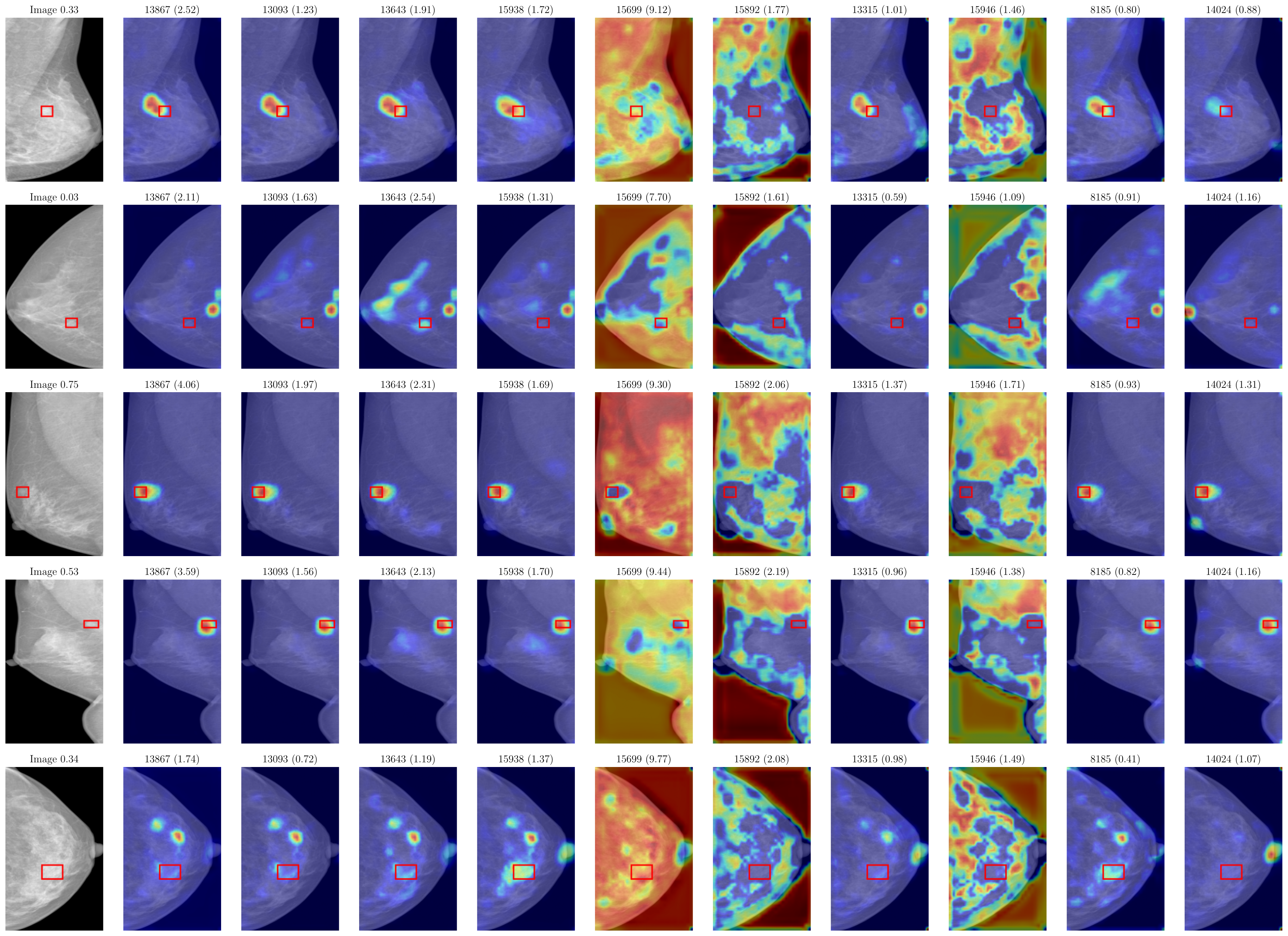}
    \caption{{Mass} (Finetuned)}
    \label{fig:topkactivatedmass}
\end{subfigure}

\caption{Visualization of the top-10 class-level latent neurons of class $c=1$ from the finetuned Mammo-SAE model for two breast cancer concepts. Red boxes denote ground-truth concept regions. Each image is annotated with the latent neuron index and its mean activation value. Best viewed in zoom. Additional examples are provided in the Appendix.}

\label{fig:topk_latent_concepts}
\end{figure}

Figures~\ref{fig:topkactivatedsc} and~\ref{fig:topkactivatedmass} show heatmaps of the top-10 latent neurons most associated with the positive class ($c=1$) for two breast cancer concepts: \textit{suspicious calcification} and \textit{mass}. Ground-truth concept regions are overlaid in red for clarity. For suspicious calcification, we observe that 7 out of the top 10 latent neurons activate strongly within the annotated region, indicating that Mammo-SAE has learned semantically meaningful and spatially aligned representations. In contrast, for the mass concept, the top-activated neurons show weak alignment with ground-truth regions, which aligns with the relatively lower AUC-ROC observed in Table~\ref{tab:saeperformace}.

To quantitatively assess the spatial alignment between SAE latent activations and annotated breast concept regions, we threshold each latent heatmap at the 95th percentile and extract rectangular bounding boxes to approximate predicted concept locations. Table~\ref{tab:concept_localization} reports the mean Average Precision (mAP) at an Intersection-over-Union (IoU) threshold of 0.25, computed over the top-10 class-selective latent neurons for both the pretrained and finetuned variants of Mammo-CLIP.
We find that the fine-tuned model consistently achieves higher mAP than the pretrained variant, suggesting that fine-tuning enhances the model’s ability to align concept-relevant features with spatially meaningful regions. Conversely, the lower mAP in the pretrained model indicates that the model often relies on spurious or task-irrelevant background regions when making predictions. It is important to note that no annotated localization information is used during training for either the pretrained or finetuned models.

These findings highlight two key insights: (1) a significant fraction of latent neurons capture clinically meaningful visual concepts, and (2) some neurons still respond to irrelevant background areas yet influence the final decision. Understanding and controlling for these background-sensitive neurons could be crucial for building more robust and interpretable breast cancer detection models. Our framework provides a concrete path forward for future efforts to mitigate reliance on confounding features during both training and inference.

\begin{table}[t]
\centering
\caption{Mean Average Precision (mAP) for breast concept localization using top-$10$ class-level latent neuron activations of class $c=1$ across different breast cancer concepts and models.}
\label{tab:concept_localization}

\resizebox{\textwidth}{!}{%
\begin{tabular}{lcccccccccccc}
\toprule
\textbf{Concept} & \textbf{Model} &  \textbf{1} &  \textbf{2} &  \textbf{3} &  \textbf{4} &  \textbf{5} &  \textbf{6} &  \textbf{7} &  \textbf{8} &  \textbf{9} &  \textbf{10}   \\
\midrule
Suspicious Calcification & Finetuned &  0.256 & 0.007 & 0.005 & 0.007 & 0.084 & 0.102 & 0.084 & 0.083 & 0.166 & \textbf{0.278} \\
Suspicious Calcification &  Pretrained & 0.057 & 0.053 & 0.027 & \textbf{0.085} & 0.008 & 0.002 & 0.002 & 0.011 & 0.033 & 0.053 \\
Mass  & Finetuned & 0.295 & \textbf{0.316} & 0.308 & 0.286 & 0.0 & 0.0 & 0.159 & 0.0 & 0.182 & 0.135 \\
Mass  & Pretrained & \textbf{0.045} & 0.019 & 0.029 & 0.053 &  0.022 & 0.020 & 0.024 & 0.022 & 0.018 & 0.025 \\
\bottomrule
\end{tabular}%
}
\end{table}

\subsection{\small Latent Neuron Separation: Finetuned vs. Pretrained Mammo-CLIP}\label{sec:latentseparationsc}

\begin{figure}[H]
    \centering
    \begin{subfigure}[t]{0.45\linewidth}
        \centering
        \includegraphics[width=\linewidth]{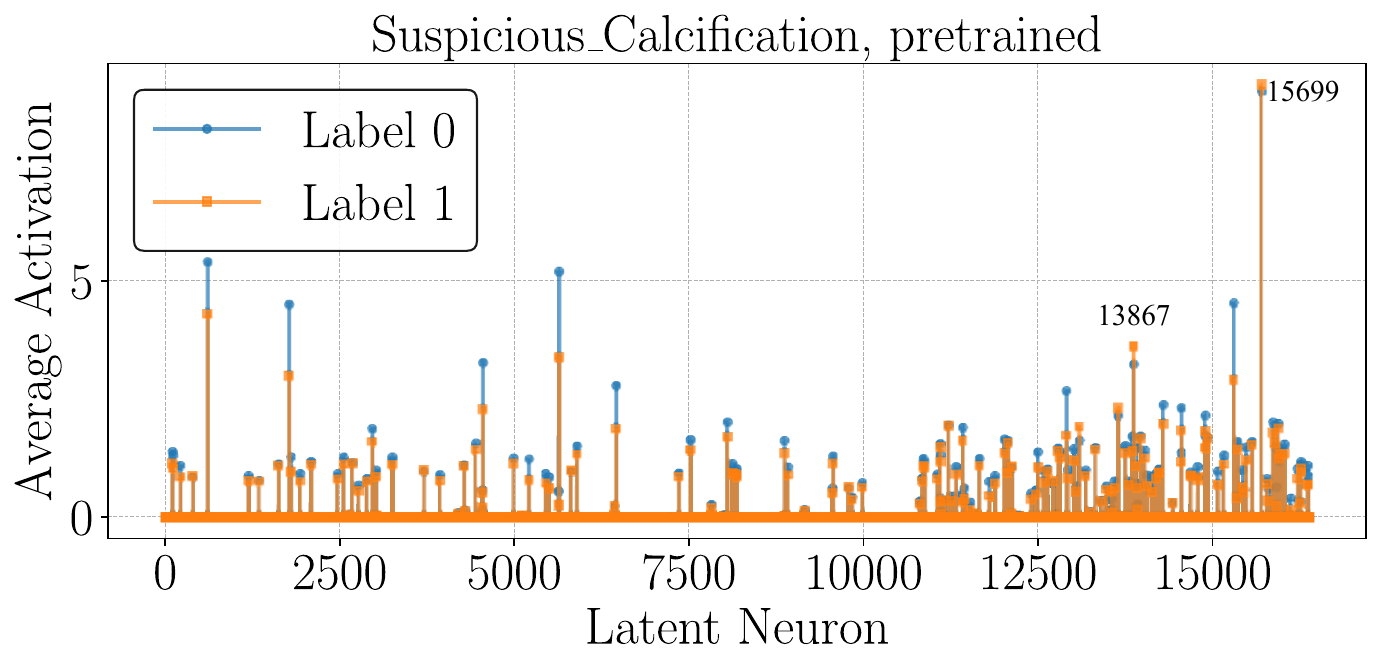}
        \caption{Pretrained}
        \label{fig:scmeanpretrained}
    \end{subfigure}
    \begin{subfigure}[t]{0.45\textwidth}
        \centering
        \includegraphics[width=\linewidth]{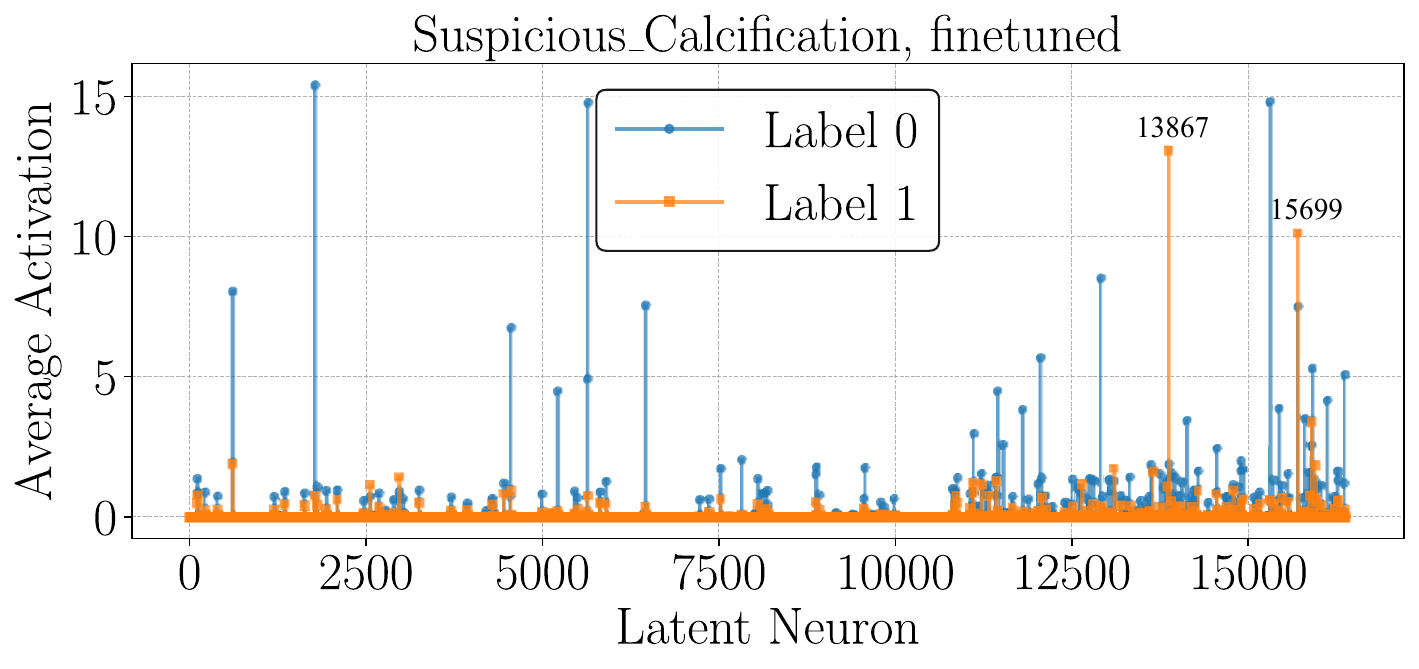}
        \caption{Finetuned}
        \label{fig:scmeanfinetuned}
    \end{subfigure}
\caption{Mean latent activation vectors $\bar{\mathbf{z}}^{(c)}$ for each class ($c=1$ indicates the presence of the concept) in the pretrained model (left) and finetuned model (right) for the \textit{suspicious calcification} concept.}
\label{fig:scfinetunedmean}
\end{figure}

Table~\ref{tab:saeperformace} demonstrates that finetuned models significantly outperform pretrained models in terms of AUC-ROC. To understand the underlying cause of this improvement, we analyze the class-wise mean latent vectors $\bar{\mathbf{z}}^{(c)}$ extracted from both models for \textit{suspicious calcification} concept, visualized in Figures~\ref{fig:scmeanpretrained} and~\ref{fig:scmeanfinetuned}.

We draw three key insights from this comparison: (1) the separation between class-wise mean activations becomes significantly more pronounced in the finetuned model, suggesting that finetuning sharpens the latent space to better distinguish the presence of breast concepts; (2) Neuron $13867$ emerges as a dominant signal for the \textit{suspicious calcification} class after finetuning which is the top-1 activated latent neuron shown in Figure~\ref{fig:topkactivatedsc} (second column), highlighting that the model learns to amplify clinically meaningful features; however (3) Neuron $15699$ remains persistently active across both classes in \textit{both} models and classes, corresponding to a spurious background region (Figure~\ref{fig:topkactivatedsc}, third column), which show the evidence that even with finetuning, the model partially relies on non-discriminative or confounding features.

\section{Conclusion}

In this paper, we introduced \texttt{\textbf{Mammo-SAE}}, a framework for uncovering breast concept representations in the Mammo-CLIP~\cite{ghosh2024mammo} foundation model. By probing the latent space, we identified neurons that are selectively activated in the presence of clinically relevant breast concepts such as \textit{mass} and \textit{suspicious calcification}. Through visualization and by concept localization, we observed that while some latent neurons align well with ground-truth regions, others respond to background areas—highlighting both the strengths and limitations of the learned representations. We believe Mammo-SAE provides a valuable tool for understanding the causal mechanisms within foundation models for medical imaging and opens new avenues for inference-time interventions to improve interpretability and performance in breast cancer detection.

\noindent {\bf Disclosure of Interests.} The author declares no competing interests relevant to the content of this article.

\section{Acknowledgements}
Author sincerely acknowledges the CHUV Breast Cancer Tumour Board team\footnote{\url{https://centrescancer.chuv.ch/equipe/rencontrer-lequipe-du-centre-du-sein/}}, including Dr. Khalil Zaman, Dr. Assia Ifticene Treboux, Dr. Wendy Jeanneret Sozzi, Prof. Patrice Mathevet, and Dr. Avdulla Krasniqi, for their compassionate care of a close family member and for the inspiration that profoundly shaped both this work and the author's personal journey.

Author also acknowledge the Mammo-CLIP~\cite{ghosh2024mammo} team for open-sourcing their code and models, as well as for providing clear documentation and instructions for the preprocessing pipeline. Lastly, Author specially thank Mrs. Vedasri Nakka for her support in creating Figure~\ref{fig:framework}.

\bibliographystyle{splncs04}
\bibliography{references}

\section{Limitations}

Our study has several limitations. First, we focus exclusively on the final layer of the Mammo-CLIP model, leaving the interpretability of earlier layers unexplored. Second, our evaluation is limited to only two breast cancer-related concepts—\textit{mass} and \textit{suspicious calcification}—and does not extend to other clinically relevant findings such as \textit{nipple retraction} or \textit{skin thickening}. Third, our analysis is confined to Mammo-CLIP; applying this interpretability framework to other vision-language models in medical imaging, such as MedCLIP~\cite{wang2022medclip} or GLoRIA~\cite{huang2021gloria}, remains an important direction for future work. Finally, our visual probing is restricted to the top-10 class-level latent neurons per class, which may overlook other relevant or informative neurons.

\section{Additional Details}\label{sec:additionaldetails}

Table~\ref{tab:model_comparison} summarizes the key differences between the \textit{pretrained} and \textit{finetuned} Mammo-CLIP~\cite{ghosh2024mammo} models in the context of breast concept prediction. While the pretrained model relies on fixed CLIP representations, the finetuned model adapts the entire backbone using supervised concept labels, potentially leading to more discriminative and task-aligned feature representations. \\

\begin{table}[h]
\centering
\caption{{Comparison of Pretrained and Finetuned Mammo-CLIP~\cite{ghosh2024mammo}}}
\label{tab:model_comparison}
\resizebox{0.95\linewidth}{!}{%
\begin{tabular}{@{}p{4cm}|p{5cm}|p{5.5cm}@{}}
\toprule
\textbf{Aspect} & \textbf{Pretrained Model} & \textbf{Finetuned Model} \\
\midrule
\textbf{Backbone Weights} & Frozen after pretraining on image--report pairs & Updated during fine-tuning \\ \hline
\textbf{Downstream Head} & Only classification head trained & Backbone and head trained jointly \\
\bottomrule
\end{tabular}%
}
\end{table}

\section{Additional Results}

\begin{figure}[H]
    \centering
    \begin{subfigure}[t]{0.45\linewidth}
        \centering
        \includegraphics[width=\linewidth]{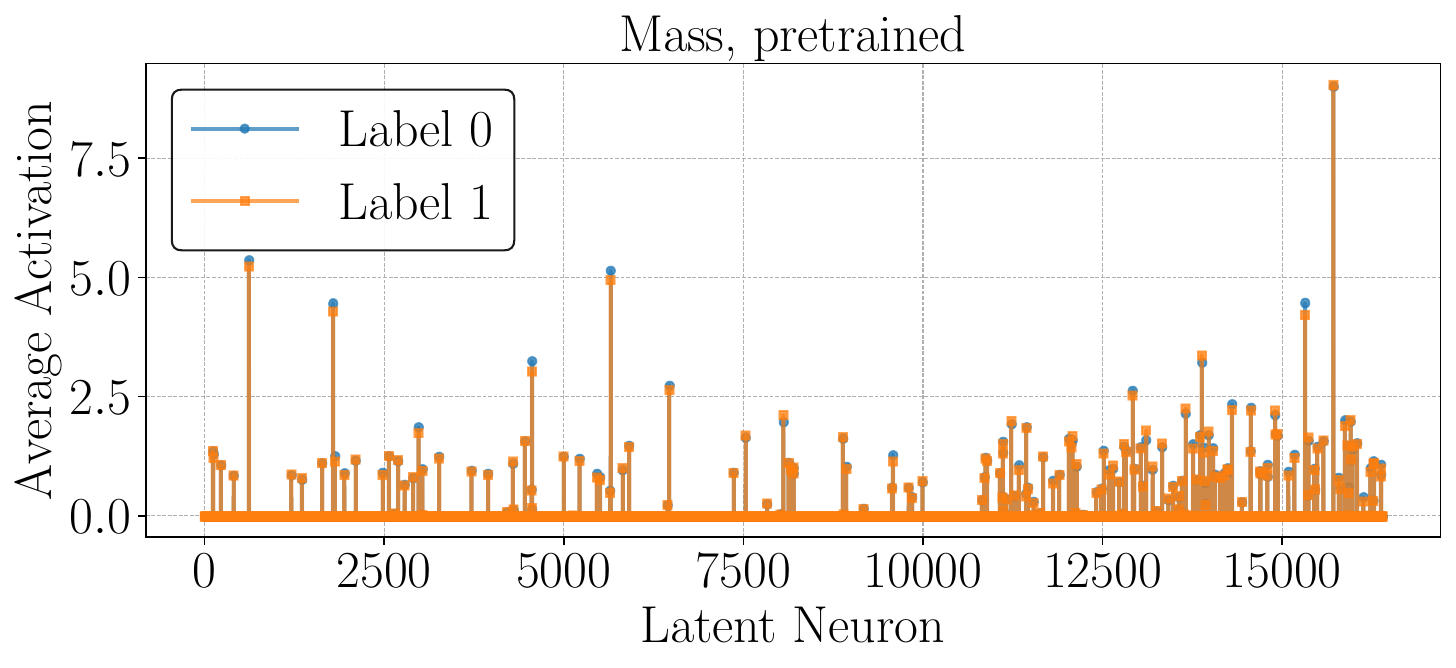}
        \caption{Pretrained}
        \label{fig:massmeanpretrained}
    \end{subfigure}
    \begin{subfigure}[t]{0.45\textwidth}
        \centering
        \includegraphics[width=\linewidth]{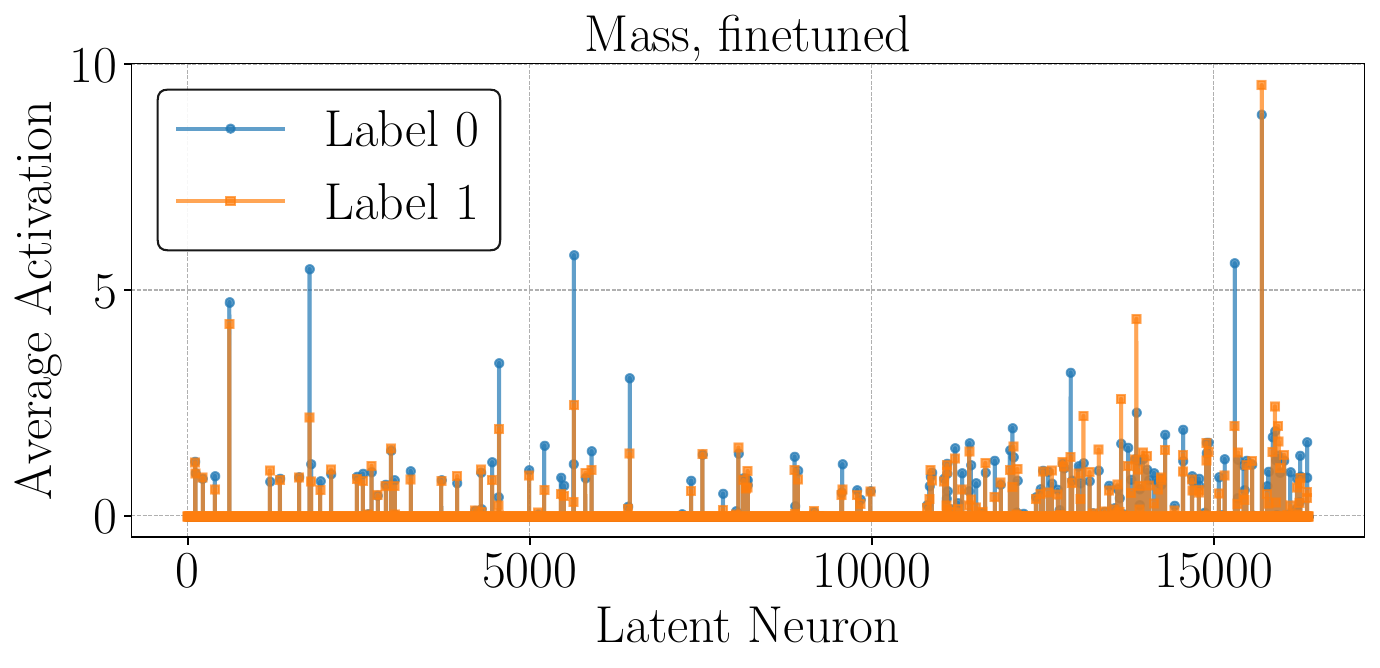}
        \caption{Finetuned}
        \label{fig:massmeanfinetuned}
    \end{subfigure}

\caption{Mean latent activation vectors $\bar{\mathbf{z}}^{(c)}$ for each class ($c=1$ indicates the presence of the concept) in the pretrained model (left) and finetuned model (right) for the \textit{mass} concept.}
\label{fig:massfinetunedmean}
\end{figure}

In Figure~\ref{fig:massfinetunedmean}, we show the mean latent activation vectors for different classes for the \textit{mass} concept using both the pretrained and finetuned models. Consistent with our observations in Section~\ref{sec:latentseparationsc}, we find that the finetuned model exhibits a more pronounced separation between the class-wise activations, suggesting improved class-specific feature encoding in the latent space.

\begin{figure}[t]
\centering
\includegraphics[width=\linewidth]{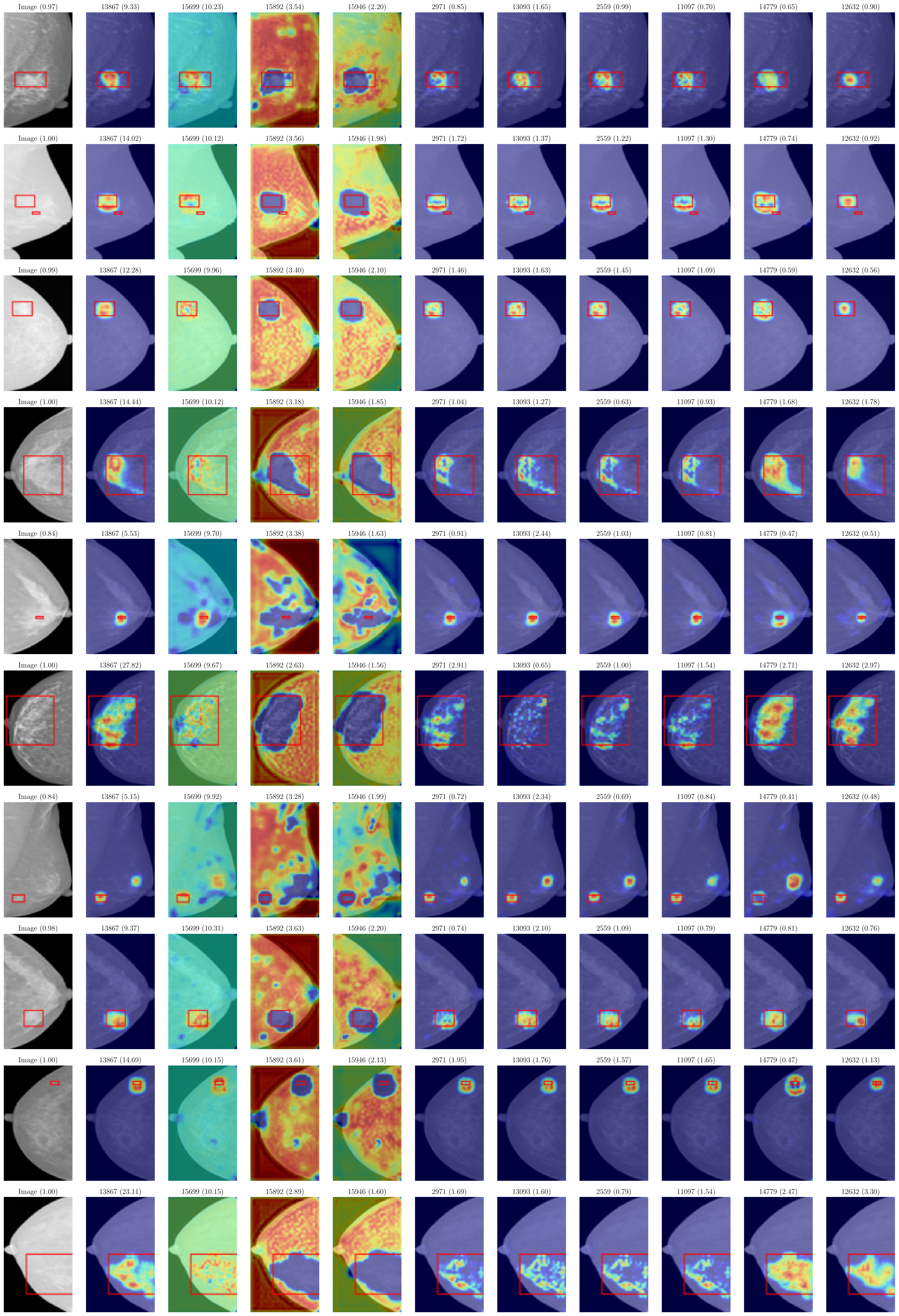}
\caption{{ Visualization of Top-$10$ Latent Neurons of class $c=1$ for \textit{Suspicious Calcification} (Finetuned Model).} Red boxes indicate ground-truth annotations.}

\label{fig:result1}
\end{figure}

\begin{figure}[t]
\centering
\includegraphics[width=\linewidth]{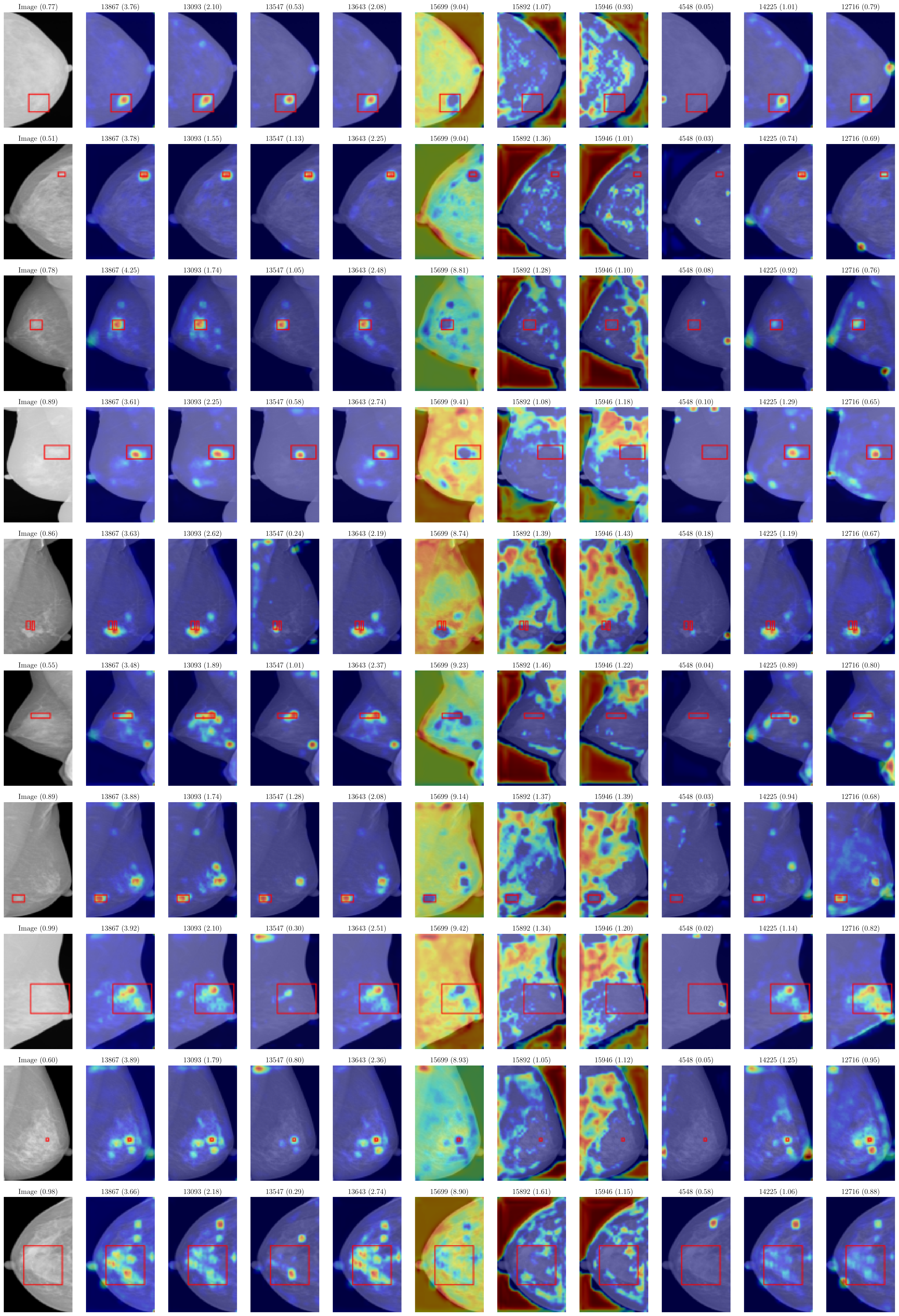}
\caption{{ Visualization of Top-$10$ Latent Neurons of class $c=1$ for \textit{Suspicious Calcification} (Pretrained Model).} Red boxes indicate ground-truth annotations.}

\label{fig:result2}
\end{figure}

\begin{figure}[t]
\centering
\includegraphics[width=\linewidth]{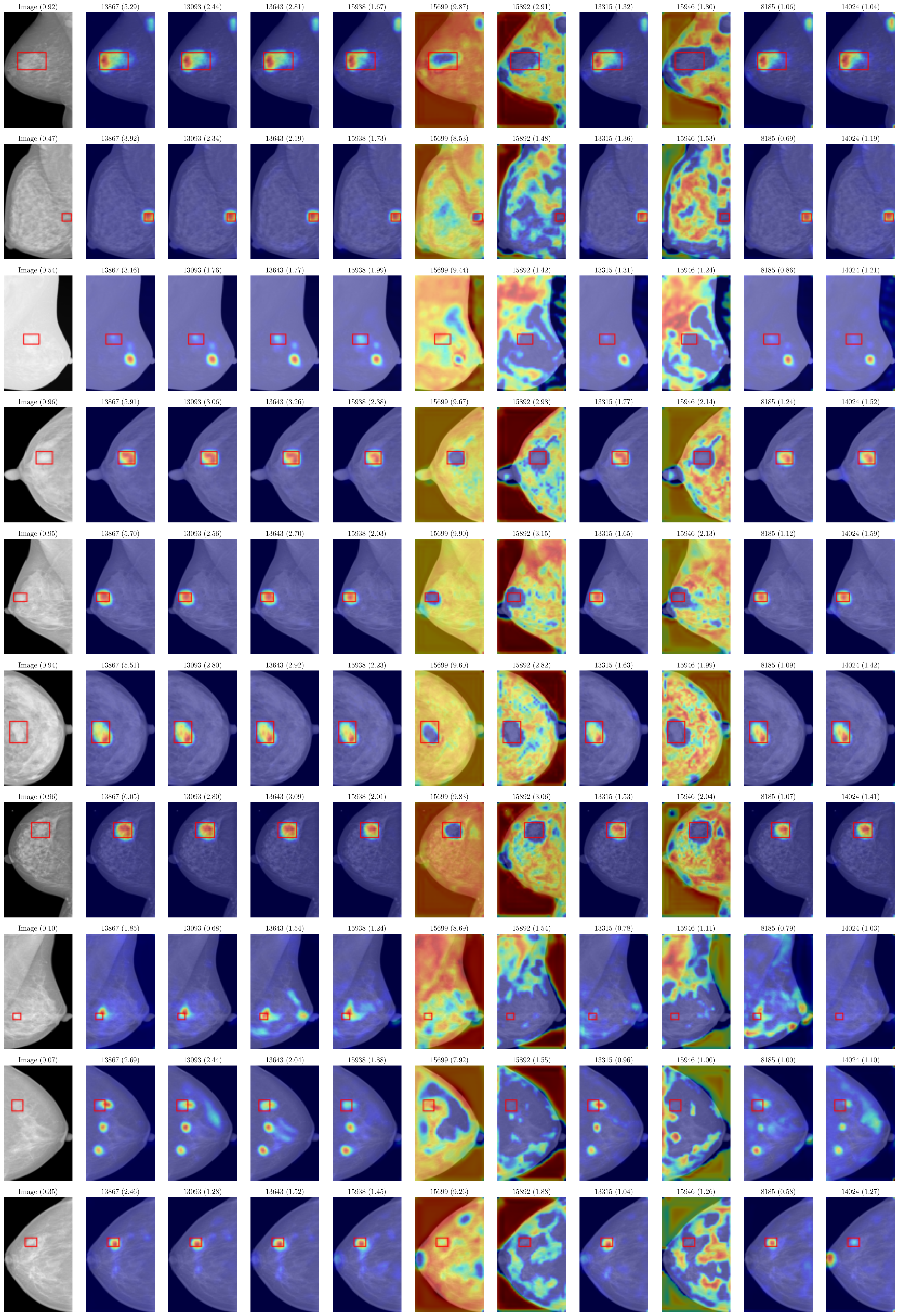}
\caption{{ Visualization of Top-$10$ Latent Neurons of class $c=1$ for \textit{Mass Calcification} (Finetuned Model).} Red boxes indicate ground-truth annotations.}

\label{fig:result3}
\end{figure}

\begin{figure}[t]
\centering
\includegraphics[width=\linewidth]{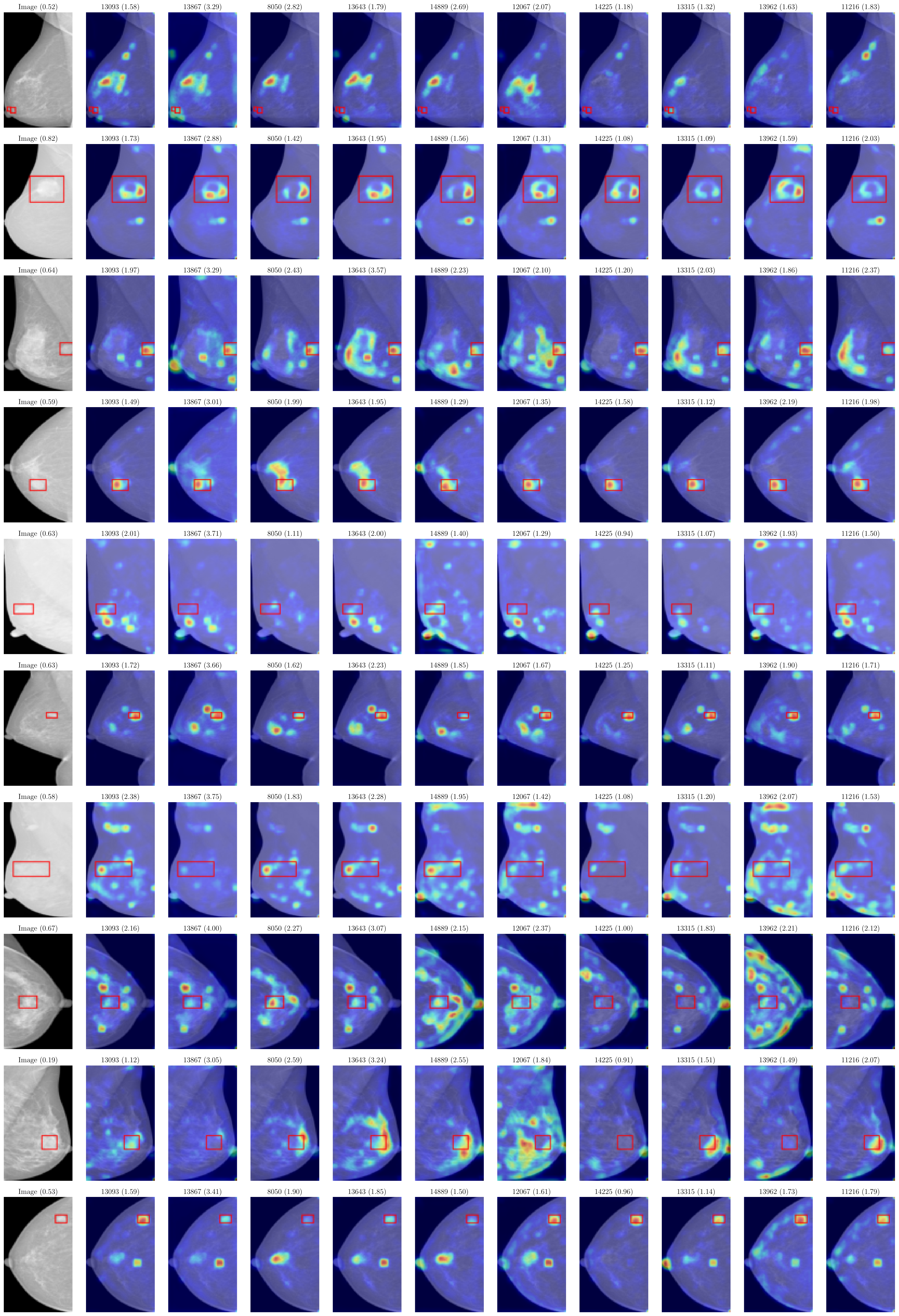}
\caption{{ Visualization of Top-$10$ Latent Neurons of class $c=1$ for \textit{Mass Calcification} (Pretrained Model).} Red boxes indicate ground-truth annotations.}

\label{fig:result4}
\end{figure}

Furthermore, in Figures~\ref{fig:result1},~\ref{fig:result2},~\ref{fig:result3}, and~\ref{fig:result4}, we present additional heatmaps of the top-10 class-level latent neurons across various model and concept combinations.

\end{document}